\documentclass[runningheads]{llncs}

\usepackage{eccv}

\usepackage{eccvabbrv}

\usepackage{epsfig}
\usepackage{url}            %
\usepackage{booktabs}       %
\usepackage{amsfonts}       %
\usepackage{nicefrac}       %
\usepackage{microtype}      %
\usepackage{xcolor}         %
\usepackage{colortbl}         %

\usepackage{graphicx}
\usepackage{blindtext}
\usepackage{amsmath}
\usepackage{amssymb}
\usepackage{bbm}
\usepackage{bm}
\usepackage{booktabs}
\usepackage{alias}
\usepackage{lib}
\usepackage{cuted}
\usepackage{xcolor}
\usepackage{array}
\usepackage{multirow}
\newcolumntype{C}[1]{>{\centering\arraybackslash}p{#1}}
\usepackage{wrapfig}
\usepackage{subcaption}
\usepackage{pifont}%
\usepackage{setspace}

\usepackage[accsupp]{axessibility}  %

\usepackage{hyperref}

\usepackage{orcidlink}

\begin{document}

\title{Mitigating Perspective Distortion-induced Shape Ambiguity in Image Crops}

\author{Aditya Prakash \and Arjun Gupta \and Saurabh Gupta}

\authorrunning{A.~Prakash et al.}

\institute{University of Illinois Urbana-Champaign\\
\email{\{adityap9,arjung2,saurabhg\}@illinois.edu}\\
\url{https://bit.ly/AmbiguityEnc}}

\maketitle

\begin{abstract}
Objects undergo varying amounts of perspective distortion as they move across a camera's field of view. Models for predicting 3D from a single image often work with crops around the object of interest and ignore the location of the object in the camera's field of view. 
We note that ignoring this location information further exaggerates the inherent ambiguity in making 3D inferences from 2D images and can prevent models from even fitting to the training data. 
To mitigate this ambiguity, we propose Intrinsics-Aware Positional Encoding (\methodname), which incorporates information about the location of crops in the image and camera intrinsics. Experiments on three popular 3D-from-a-single-image benchmarks: depth prediction on NYU, 3D object detection on KITTI \& nuScenes, and predicting 3D shapes of articulated objects on ARCTIC, show the benefits of \methodname.
\end{abstract}

\section{Introduction}
\label{sec:intro}
Making metric predictions from a single image suffers from scale-depth ambiguity. Even when the camera intrinsics are known, it is impossible to disambiguate if Circle A shown in \Figref{circle} has a radius of 0.485 cm (say) \& is 2 cm away (say) or a radius of 4.85 cm and 20 cm away. Familiar size, or knowledge of the size of a reference object in the image, is a useful cue for resolving this ambiguity~\cite{epstein1961known, ittelson1951size}.

The scale-depth ambiguity is a well-known and fundamental ambiguity. What we describe next is a subtle different ambiguity that occurs when we consider object crops from images for feeding into a neural network. Let's assume that we are using size familiarity to resolve scale-depth ambiguity, \ie assume that the circles always have a radius of 0.485 cm. Given this information, it should be possible to infer how far they are in the image. This can be readily derived using similar triangles and the {\it location of the circle's image in the visual field}. In fact, this principle goes back many centuries and was known to Da Vinci~\cite{vinci1632treatise}.

However, if we consider crops around the object, say those output by an object detector, and try making a prediction for the circle's distance based on the appearance of the crop, this can prove to be quite difficult. \Figref{circle} shows 
\setlength{\intextsep}{0pt}%
\begin{wrapfigure}[32]{r}{0.50\textwidth}
\includegraphics[width=1.0\linewidth,page=1]{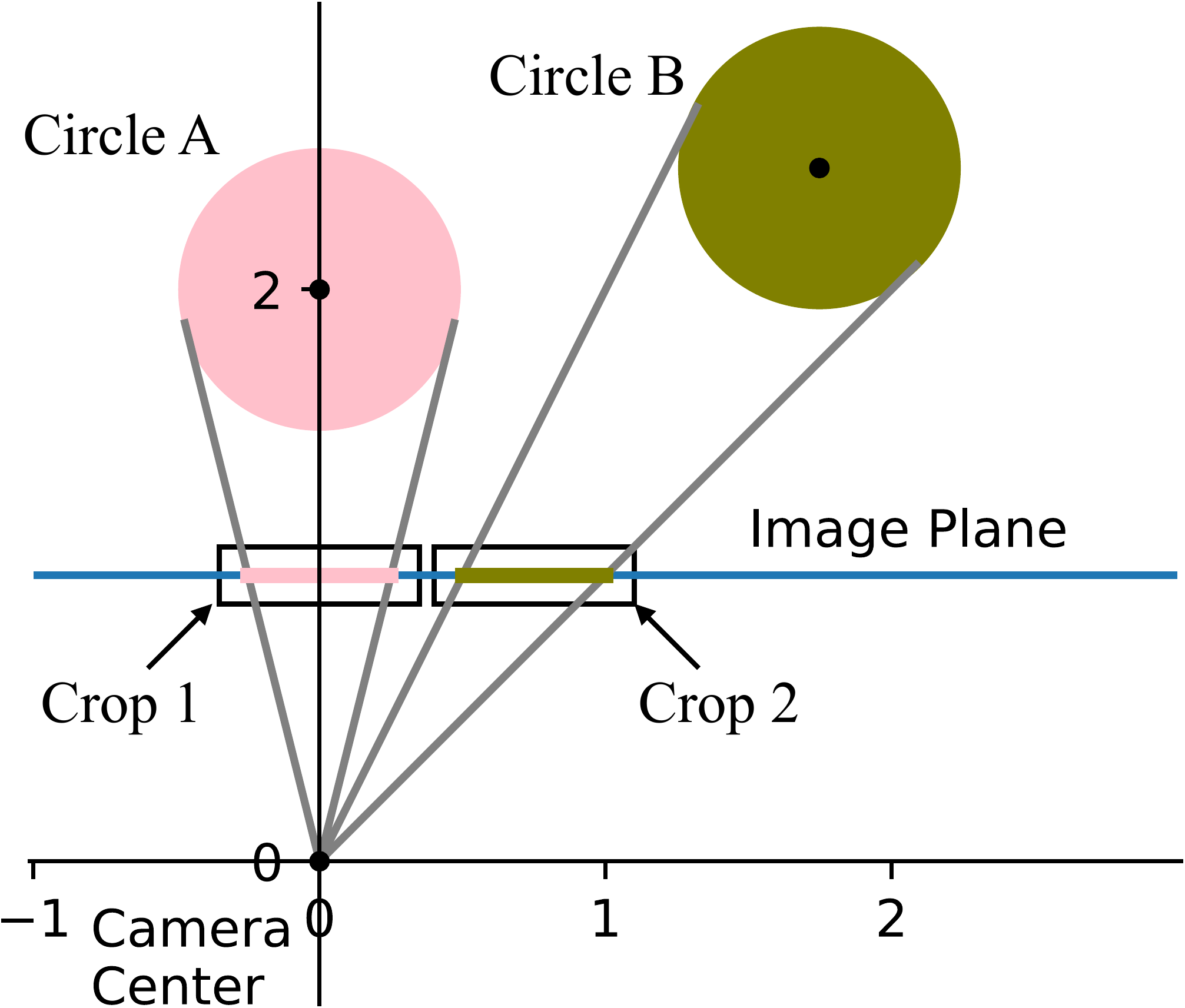}
\caption{\textbf{Perspective Distortion-induced Shape Ambiguity}. Consider two circles of the same size undergoing perspective projection under a pinhole camera. Even though they are at different distances from the camera, they appear to be the same size in the image due to perspective distortion. A model (e.g. a neural network) that predicts the distances of these circles from the camera based purely on the appearance of the image crops, {\it without taking into account their location in the camera's field of view}, will fail at this task. We call this the {\it Perspective Distortion-induced Shape Ambiguity in Image Crops} or {\it \name}. In this work, we propose an encoding to incorporate the crop location in the camera's field of view as input and show its effectiveness on metric depth prediction, 3D object detection \& 3D pose estimation of articulated objects (\Secref{experiments}).} 
\figlabel{circle}
\end{wrapfigure}%
fixed sized center crops around the center of the circle's image. It turns out, just looking at the crops while ignoring the location of the crop, it is no longer
possible to predict the distance of the circle from the camera! A 0.5 cm wide image could correspond to Circle A that is 2 cm away directly in front of the camera, or Circle B that is further away (2.43 cm away) but off to the side. This simple example only exposed global pose error, but our detailed case study in \Secref{toy} reveals that this ambiguity exists both in global pose and root-relative 3D shape.  We call this {\it Perspective Distortion-induced Shape Ambiguity in Image Crops or \name}.

To mitigate this ambiguity, we propose a intrinsics-aware positional encoding (\methodname) that incorporates the location of the crop in the camera's field of view (\Figref{method}). This allows the network to specialize its interpretation of 3D geometry based on the location in the image. We show results on three popular 3D-from-single-image benchmarks: articulated object pose estimation on \arctic~\cite{Fan2023CVPR}, pixel-wise metric depth predictions on \nyu~\cite{silberman2012indoor} and 3D object detection from monocular images on \kitti~\cite{geiger2012we} and nuScenes~\cite{Caesar2020CVPR}. Across these three real world benchmarks and a diagnostic setting, we find this additional information to improve metrics. 

We acknowledge that similar encodings have been used in literature~\cite{guizilini2023towards, facil2019cam}. However, their use was motivated by the need to train models across different cameras. Our contribution is to identify perspective distortion-induced shape ambiguity in image crops, even with a single camera, and show the effectiveness of encoding the location of the crop in the camera's field of view to mitigate this ambiguity.

\begin{figure}[t]
\resizebox{\linewidth}{!}{
\begin{tabular}{ccc}
\insertH{0.2}{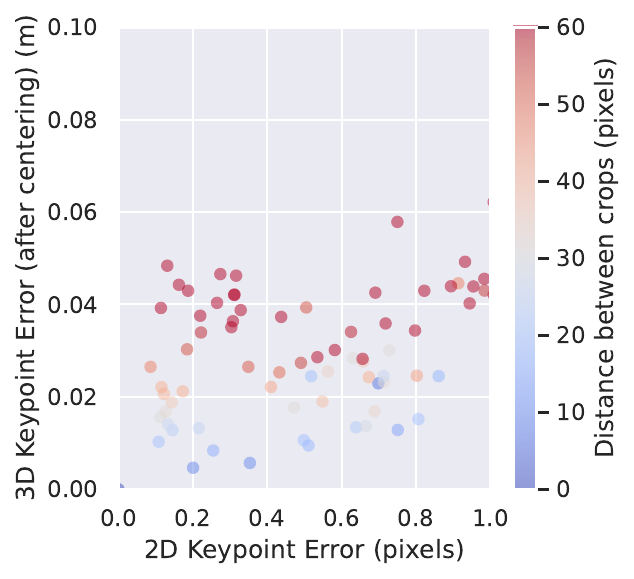} &
\insertH{0.2}{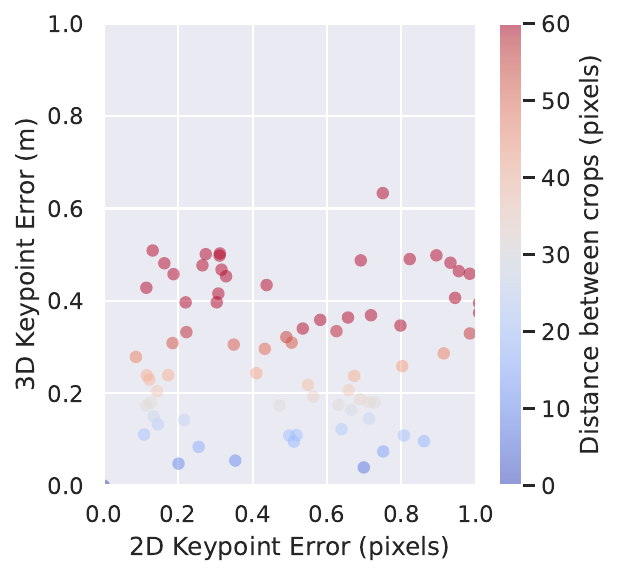} &
\begin{tikzpicture}
  \node[inner sep=0pt] (vis) at (0,0)
  {\includegraphics[height=.2\textwidth]{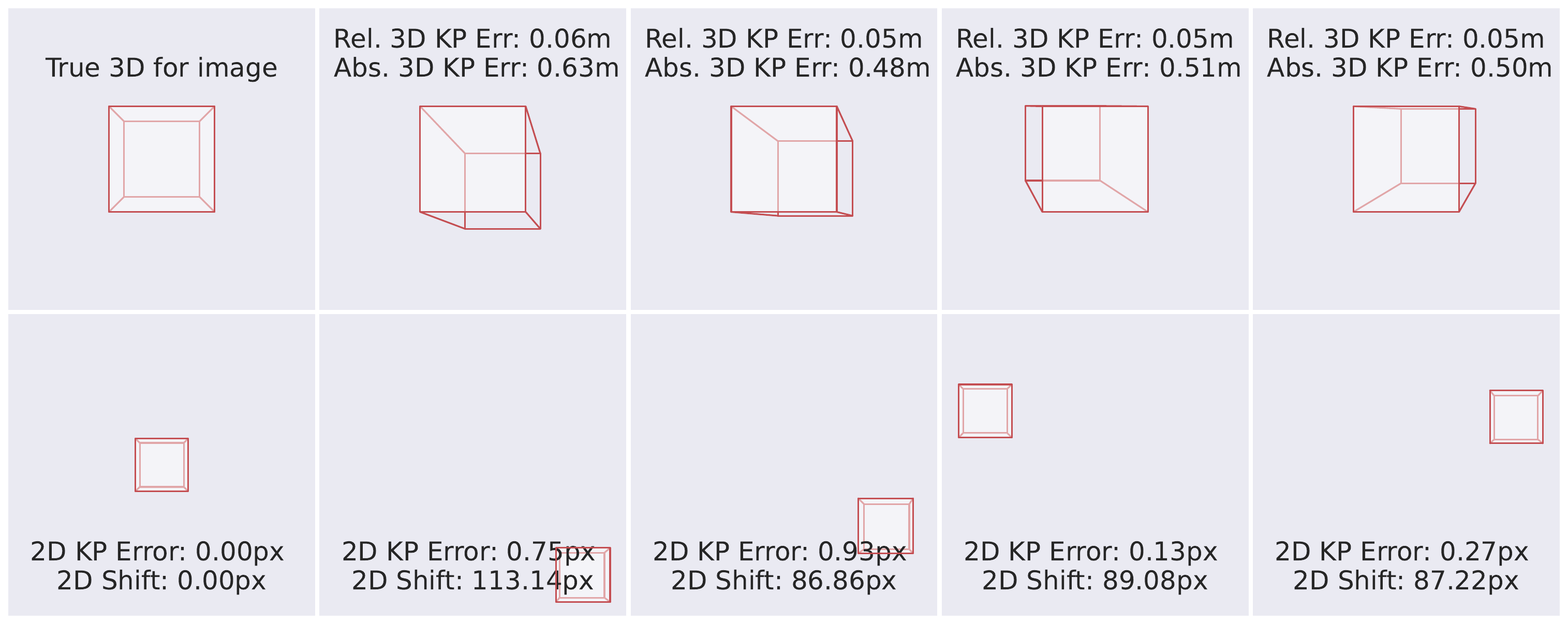}};
  \draw[darkgreen, very thick] (-3.029,-1.1875) rectangle ++(1.1875,2.375);
\end{tikzpicture} \\
\scriptsize (a) Root-relative 3D & \scriptsize (b) Absolute 3D &
\scriptsize (c) Example Ambiguous Parallelepipeds. \\ %
\end{tabular}
}
\caption{\textbf{Parallelepipeds Case Study}. 
Figure (a) plots the root relative 3D keypoint error \vs 2D keypoint
error of different parallelepipeds placed at different locations in the
camera's field of view \wrt a reference cuboid shown in the {\color{darkgreen}
green} box in (a.2).
Points are color coded with the distance between the parallelepipeds crops. As
we let crops go farther away (the {\color{red}red} points), we start finding parallelepipeds
that have very different 3D shape but happen to project such that their 2D
keypoints look the same as the reference cuboid. Figure (b) shows a similar
plot but for absolute 3D keypoint error. Figure (c) shows these ambiguous 3D
parallelepipeds in the top row and their renderings in the bottom row. The
1\textsuperscript{st} figure in the {\color{darkgreen} green} box is the
reference \wrt which we measure 2D and 3D keypoint errors.}
\figlabel{parallelepipeds}
\end{figure}

\section{Related Work}
\label{sec:related}

\boldparagraph{Ambiguity in 3D from a single image}
Scale-depth ambiguity is well known and is commonly dealt with by only evaluating 3D predictions up to a scale~\cite{Ranftl2022PAMI} or after a rigid alignment with the ground truth~\cite{kanazawa2018end}. Where metric 3D predictions are needed, training is either done using domain specific knowledge~\cite{bhat2023zoedepth}, camera intrinsics as input~\cite{brazil2023omni3d} or predicted camera calibration parameters~\cite{jin2023perspective, kar2015amodal}, \eg focal length~\cite{kar2015amodal} using appearance cues~\cite{jin2023perspective}. Notably, due to this ambiguity, models can not trivially be trained  with scale and crop augmentations unless some adjustments are made~\cite{brazil2023omni3d,eigen2014depth}. Ambiguity induced due to crops (the focus of our work) has not been analyzed in literature to the best of our knowledge. 
Note that this ambiguity is an artifact of {\it planar} perspective projection in images and doesn't exist in {\it spherical} perspective projection. The relationship between planar and spherical perspective have been used to simplify analysis, \eg \cite{malik1997computing}.

\boldparagraph{Geometric embeddings} 
Past works have considered the use of different encodings, based on pixel
locations~\cite{liu2018intriguing, dosovitskiy2020image}, camera
intrinsics~\cite{facil2019cam} or extrinsics~\cite{zhao2021camera,
yifan2022input}, as input to networks to improve performance. Liu~\etal
\cite{liu2018intriguing} show that preserving spatial location of the pixels
helps in generative modelling, object detection, and reinforcement learning.
Different forms of positional encodings, \eg learnable~\cite{dosovitskiy2020image,Karras2019CVPR}, cartesian spatial grid~\cite{Jaderberg2015NEURIPS}, sinusoidal~\cite{Vaswani2017AttentionIA,Mildenhall2020ECCV}, implicit~\cite{Xu2021CVPR,Shaham2019ICCV}, help incorporate spatial inductive biases in GANs~\cite{Xu2021CVPR}. 
Positional encodings, \eg learned~\cite{dosovitskiy2020image},
relative~\cite{sun2022alength}, or rotational~\cite{su2021roformer}, are also
commonly use in training transformer~\cite{Vaswani2017AttentionIA}-based
models. Recent works~\cite{facil2019cam, guizilini2023towards} have also
studied the effectiveness of pixel-level encodings in multi-dataset setting
(with varying cameras) on depth estimation tasks. Researchers have also
considered canonicalizing the inputs (captured with different cameras) to a
common camera intrinsics~\cite{brazil2023omni3d, antequera2020mapillary} to
indirectly provide camera information. \cite{yifan2022input,
guizilini2022depth, miyato2023gta} employ this idea in multi-view settings.
Yifan \etal \cite{yifan2022input} used such encodings to inject geometric
inductive biases (\eg the epipolar geometry constraint) into a general
perception model for multi-view stereo. Guizilini \etal
\cite{guizilini2022depth} extend this further and use per-pixel embeddings to
capture the image coordinate and camera intrinsics \& extrinsics for multi-view
depth estimation. Miyato \etal \cite{miyato2023gta} propose a geometry-aware
attention mechanism for transformers for novel view synthesis tasks.

Our use of the location of the crop in the camera's field of view as an additional input to neural networks is motivated by these past works. Orthogonal to these works, our contribution is to demonstrate the effectiveness of this encoding in mitigating the perspective distortion-induced shape ambiguity in image crops. We conduct this study across 3 representative 3D from a single image tasks, and where appropriate, show comparisons to past embeddings: Cam-Conv~\cite{facil2019cam} and sinusoidal image location embeddings~\cite{Vaswani2017AttentionIA}.

\section{Parallelepipeds Case Study}
\seclabel{toy}

We first analyze simple parallelepipeds to intuitively understand the perspective distortion-induced ambiguity in image crops and the effectiveness of the crop location in the camera's field of view to mitigate this ambiguity. It is important to note that this ambiguity does not always exist. If the shape and size of the object are perfectly known (say we know that it is a front-facing cube), then the location of the cube in the visual field can itself be deduced from an image crop.

However, if the geometry of the object is also unknown, then it may no longer
be possible to disentangle shape \& pose from a given image crop. Consider the
family of 3D shapes with a square face (of width $0.2$m lying in XY plane) that
is extruded along an arbitrary direction, \ie parallelepipeds with atleast one
face being a square. \Figref{parallelepipeds}\,(c, top row) shows different
parallelepipeds when placed directly in front of the camera (\ie from one
consistent viewing angle across the row). Note that the parallelepipeds here do
not extend purely in the Z-direction (except for the leftmost reference in
green box), but can extend at an oblique angle. \Figref{parallelepipeds}\,(c bottom row) shows these same parallelepipeds when placed at different locations in the visual field. These locations were mined separately for each so as to maximize the visual similarity to the leftmost reference parallelepiped. As evident, even though the parallelepipeds across the row have different shapes, they all look the same in the 2D image (with a sub-pixel reprojection error), albeit for their 2D locations.

\begin{figure}[t]
    \centering
    \insertWL{0.8}{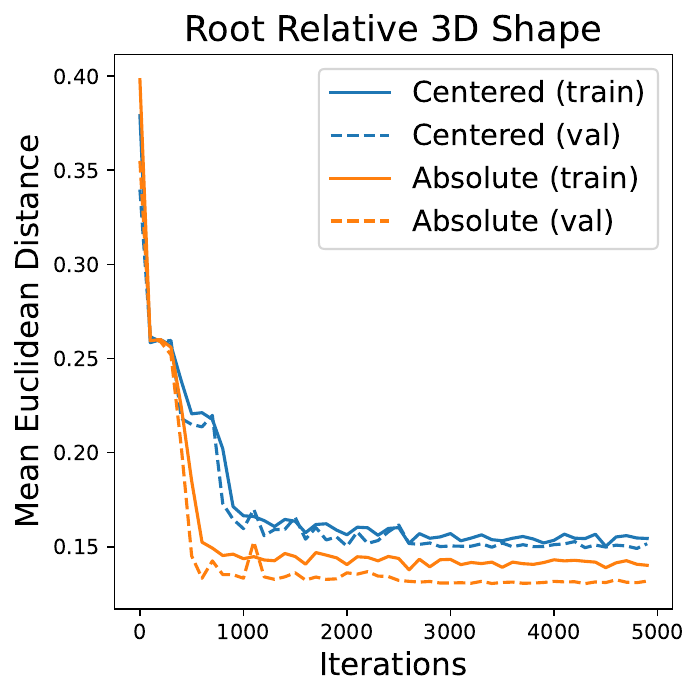}
    \insertWL{0.8}{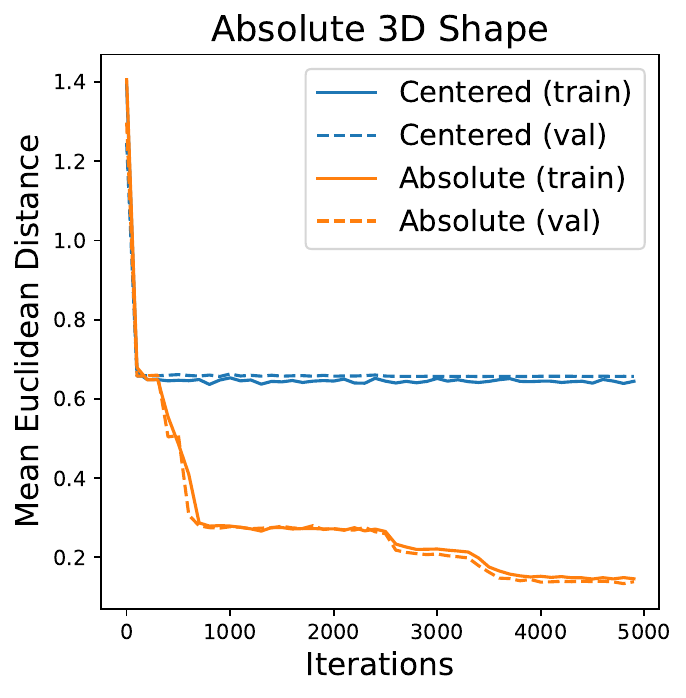}
    \caption{Predicting root relative (left) or absolute (right) 3D shape from
    2D image crops fails in the absence of information about the location of
    the crop in camera's field of view. Training loss saturates at a high value
    because of the inherent ambiguity. Adding information about the location of
    the crop in camera's field of view alleviates this ambiguity, leading to
    better metrics for both root-relative and absolute 3D prediction.}
    \figlabel{parallelepipeds-network}
\end{figure}

This ambiguity only exists if we only look at the image crops locally without
factoring in the absolute 2D location of the crop. Let's treat the
parallelepiped in the first column (a cuboid) as reference.
\Figref{parallelepipeds}\,(a) plots the root relative 3D shape error as a
function of the root relative 2D keypoint error of different parallelepipeds
\wrt the reference cuboid. For root relative error, we compute the keypoint
error \textit{after} the centers of the parallelepipeds have been aligned.
Points are color coded based on the 2D distance of the image crops from the
cuboid crop (blue means image crops are closer, red means they are farther
away). If we limit reasoning to only close-by crops, \ie the blue points, low
2D keypoint error implies low 3D shape error. However, if we look at the
farther away points (the red points), image with a low relative 2D keypoint
error can actually have a high 3D shape error. This is the inherent ambiguity:
even if two shapes appear similar visually, their 3D shapes may be vastly
different. \Figref{parallelepipeds}\,(b) presents the analogous plot where we
measure absolute 3D error instead of root-relative error.

\begin{figure*}[t]
    \centering
    \insertW{1.0}{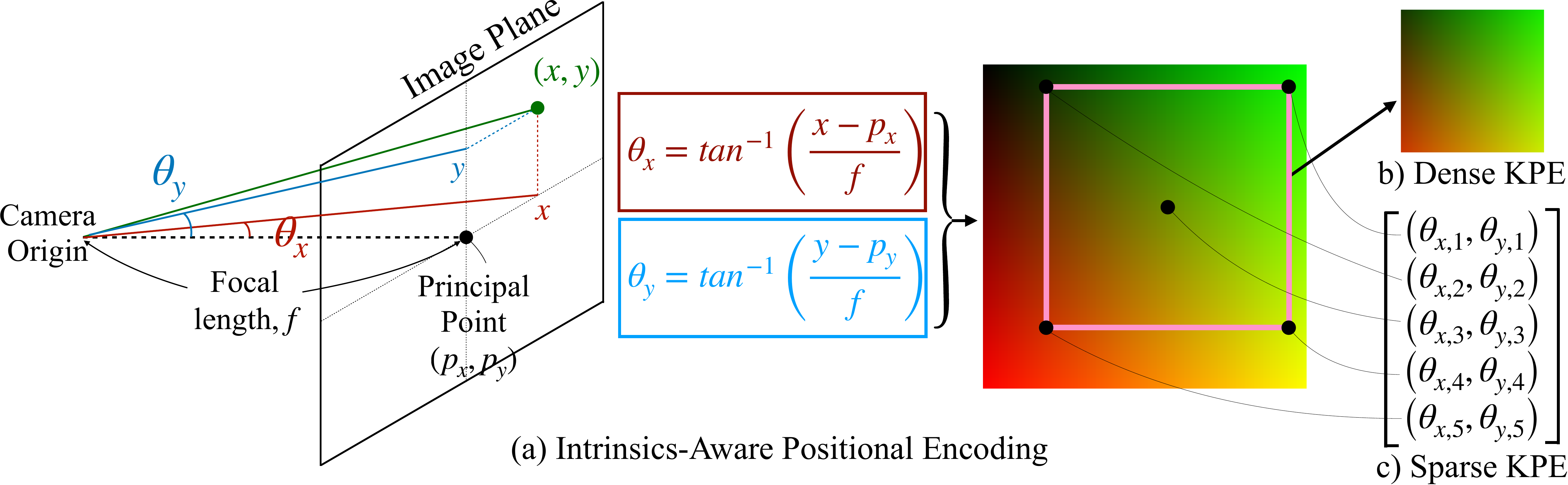}
    \caption{\textbf{Intrinsics-Aware Positional Encodings (\methodname)}. \textbf{(a)} 
    For each pixel in the image, we compute its position in the camera's field of view
    ($\theta_x$ and $\theta_y$), or the angular
    distance that the pixel makes with respect to the principal point and the camera 
    origin (as shown on the left). Note that both $\theta_x$ 
    and $\theta_y$ are sensitive to the camera intrinsic parameters. \textbf{(b)} For a dense 
    prediction task, we make use of a dense prediction encoding which contains 
    the positional encoding for each pixel in the region of interest. 
    \textbf{(c)} For other tasks, we simply represent the positions of the corners of the 
    relevant region of interest in addition to the center point. The positional
    encoding can be passed into the network at the input level or concatenated
    to some intermediate representation, this design choice is made separately
    for each task.
    }
    \figlabel{method}
\end{figure*}

Let's see what happens when we train neural networks to predict 3D shape from
the 2D keypoints. Given the 2D coordinates for the 8 corners, the goal is to
predict the root-relative (or absolute) 3D location of the 8 corners. We
consider two variants. First, we center the eight 2D corners by subtracting
their mean to get centered crops. Next, we retain the absolute value of the 2D
coordinates. We train a 6-layer MLP with these two inputs.
\Figref{parallelepipeds-network} shows the training and validation plots for
predicting root relative and absolute 3D coordinates for the 8 corners. As
expected the first version that only uses the crops converges to a higher error
even on the training set as it is impossible to infer the 3D shape from
centered 2D corners due to the ambiguity. The second version that has access to
the absolute coordinates does not suffer from this ambiguity and is
able to successfully estimate the 3D shape on both training and validation data.

\section{Intrinsics-Aware Positional Encoding (\methodname)}
\seclabel{method}

We mitigate the ambiguity induced by perspective distortion (\Figref{circle})
through design of encodings that capture the location of the crop in the
camera's field of view, referred to as Intrinsics-Aware Positional Encodings
(\methodname). Specifically, let's denote the principal point for the original
image by $(p_x, p_y)$ and focal length by $(f_x, f_y)$. For any pixel $(x,y)$,
we can describe its position in the camera's field of view via $\theta_x =
\tan^{-1} \left( \frac {x - p_x}{f_x} \right)$ and $\theta_y = \tan^{-1} \left(
\frac {y - p_y}{f_y} \right)$, as shown in \figref{method}\,(a). Our proposal
is to retain this information for pixels when images are cropped or resized for
feeding into neural networks.

\subsection{Using \camenc}

\methodname is quite flexible and can be used in different ways. We specifically consider two variations: Dense \camenc (\figref{method}\,b) and Sparse \camenc (\figref{method}\,c). 

\boldparagraph{Dense \camenc} We compute $\theta_x$ and $\theta_y$ for each per-pixel in
a dense manner. Rather than using the raw angles, we encode them via sinusoidal
encodings (4 frequencies each)~\cite{Vaswani2017AttentionIA}. The spatial feature map can be input into the network along side the raw pixels. Note that dense \methodname can be computed at different resolutions
and can be concatenated with a spatial feature maps at different points in the
network. 

\boldparagraph{Sparse \camenc} We compute $\theta_x$ and $\theta_y$ for the 4
corners and the center of the crop. As for dense \camenc, we encode them via
sinusoids. Resulting embeddings can be concatenated and fed into MLP layers
(along with flattened features for crop) or can be concatenated with spatial
feature maps via broadcasting.

We experiment with sparse and dense KPE for the 3 representative tasks and find
that dense \camenc works best for dense tasks, while sparse \camenc works best
for tasks that involve summarizing the geometric properties for objects, \eg
predicting its 3D bounding box or 3D pose.

\section{Experiments}
\seclabel{experiments}
We consider 3 challenging settings, involving the prediction of 3D geometric
attributes from a single image, to show the effectiveness of \camenc: 3D
pose estimation of articulated objects in contact (\secref{sec:arctic}), dense depth
prediction(\secref{sec:depth}) and 3D object detection
(\secref{sec:detection}). 
Since our goal is to show the utility of
\camenc (not to achieve state-of-the-art results) we select some
representative models from recent works on each of these tasks and incorporate
\camenc into their architectures. 
These tasks encompass sparse and dense prediction of 3D attributes, and the
representative methods cover both convolutional and transformer architectures.
Thus, these together comprehensively test KPE. 
In each of the following subsections, we describe the application (the
representative method and relevant metrics), how we apply \camenc, and the
results we obtain. We summarize the key takeaways in \secref{discussion}.

\subsection{Application 1: 3D Pose of Articulated Objects in Contact~\cite{Fan2023CVPR}}
\seclabel{sec:arctic}
\boldparagraph{Task}
Given an egocentric image showing an articulated object in interaction, the
goal is to estimate the 3D pose of the object in contact. Specifically, the 3D
pose is represented using rotation in the camera coordinate frame, the
translation of the root of the object with respect to the camera center and the
angle of articulation between the bottom and top components of the articulated
object.  The object mesh is assumed to be available. We also consider contact
estimation and 4D motion reconstruction to show the versatility of \camenc.

\boldparagraph{Method}
We adopt the single-image ArcticNet-SF~\cite{Fan2023CVPR} architecture released
with the ARCTIC benchmark~\cite{Fan2023CVPR}. The network takes a single image
as input, which is processed by a ResNet50~\cite{He2016CVPR} backbone to get a
feature map of resolution $7 \times 7 \times 2048$. These features are then
average pooled and passed to a HMR~\cite{kanazawa2018end}-style decoder to
predict the 7-dimensional pose, consisting of rotation, translation \&
articulated angle of the object. The network outputs rotation in axis-angle format and uses the weak perspective camera model to estimate the translation~\cite{Boukhayma2019CVPR,kanazawa2018end,Kocabas2021ICCV,zhang2019end}. The entire network is trained end-to-end using an L2 loss between the predictions and the ground truth for 7-dimensional pose, 3D keypoints and 2D projection of 3D keypoints. %

\boldparagraph{Modifications} We modify the architecture to also take crops around the object as input, which are processed by a ResNet50 backbone~\cite{He2017ICCV}. Since our network requires object crops, we also train a bounding box predictor model by fine-tuning MaskRCNN~\cite{He2017ICCV} on the ARCTIC training set using the ground truth bounding box computed from the 2D projections of 3D object keypoints. This is required to evaluate our model on the online leaderboard for which ground truth is not available. For contact prediction, we use the hand pose from the default ArcticNet-SF decoder.

\boldparagraph{Incorporating \camenc} As described in~\Secref{method}, we explore both sparse (captures only the center and corner pixels of the object crop) and dense (encode each pixel in the crop). We consider 3 choices for where to add \camenc in the model: (1) concatenate with the input, (2) process the encoding with a MLP \& add with the average pooled global features, (3) interpolate the encoding to $7 \times 7$ resolution \& concatenate with $7 \times 7 \times 2048$ feature map from the last convolutional layer of ResNet50. This concatenated feature is then processed by three convolutional layers and flattened to $2048$ dimensions before being passed to the decoder. We do not use batchnorm in these 3 convolutional layers.%

\boldparagraph{Dataset}
ARCTIC~\cite{Fan2023CVPR} is a recent dataset consisting of hands interacting with articulated objects in a free-form manner. There are 2 settings in the dataset: 1.5M frames from 8 allocentric views \& 200K frames from 1 egocentric view. We show results in egocentric setting since the perspective distortion is more prominent due to objects being closer to the camera. The released dataset does not contain ground truth for test set, which is evaluated separately on the online leaderboard. We compare with ArcticNet-SF in this setting.

\begin{table}[t]
    \centering
    \captionof{table}{\textbf{3D pose estimation of articulated objects in
    contact -- Main Result}. We compare with ArcticNet-SF~\cite{Fan2023CVPR} on multiple
    geometric tasks: 3D pose, contact and 4D motion estimation for articulated
    objects on ARCTIC~\cite{Fan2023CVPR}. Our intrinsics-aware positional encoding leads to significant improvements
    across all metrics and compares favorably to other ways of encoding the
    location of the crop and past encodings~\cite{facil2019cam}.}
    \tablelabel{tab:arctic_main}
    \setlength{\tabcolsep}{4pt}
    \resizebox{\linewidth}{!}{
    \begin{tabular}{l c c c c c c}
        \toprule
        \multirow{2}{*}{\bf Method} & \multicolumn{2}{c}{\bf Object} &
        \multicolumn{2}{c}{\bf Contact} & \multicolumn{2}{c}{\bf Motion} \\
        \cmidrule(lr){2-3} \cmidrule(lr){4-5} \cmidrule(lr){6-7}
        & AAE & Success & CDev$_{ho}$ & MRRPE$_{ro}$ & MDev$_{ho}$ & Acc \\
        & ($^\circ$) $\downarrow$  & (\%) $\uparrow$ & ($mm$) $\downarrow$ & ($mm$) $\downarrow$ & ($mm$) $\downarrow$ & ($m/s^{2}$) $\downarrow$ \\
        \midrule
        \multicolumn{7}{l}{\it a) Validation Split (\camenc helps w.r.t. other choices)} \\ 
        \quad ArcticNet-SF (full image)~\cite{Fan2023CVPR} & 8.0     & 59.0     & 44.1     & 36.8     & 11.8    & 11.3 \\
        \quad ArcticNet-SF (crop)                          & 7.0     & 66.0     & 74.3     & 63.0     & 19.3    & 10.9 \\
        \quad Full Image with Binary Mask                  & 6.8     & 66.8     & 75.6     & 60.9     & 17.8    & 10.8 \\
        \quad Cam-Conv~\cite{facil2019cam}                 & 6.3     & 68.8     & 38.7     & \bf 29.7 & 10.1    & 9.4 \\
        \quad Image Location (dense)                       & 6.1     & 69.2     & 40.4     & 31.6     & 10.1    & 9.2 \\
        \quad Image Location (sparse)                      & 6.2     & 67.9     & \bf 39.1 & 30.0     & 10.1    & 9.3 \\
        \quad \methodname (sparse)                         & \bf 5.9 & \bf 71.5 & 39.4     & \bf 29.7 & \bf 9.3 & \bf 8.7 \\
        \midrule
        \multicolumn{7}{l}{\it b) Test Split} \\
        \quad ArcticNet-SF (full image)~\cite{Fan2023CVPR} & 6.4          & 53.9          & 44.7          & 36.2          & 11.8          & 9.1 \\
        \quad \methodname (sparse)                         &
        \textbf{5.2} \gn{-19} & \textbf{64.1} \gn{+19} & \textbf{36.5} \gn{-18} &
        \textbf{28.3} \gn{-22} & \textbf{10.5} \gn{-11} & \textbf{7.6} \gn{-16.4} \\
        \bottomrule
    \end{tabular}}
\end{table}

\begin{table}[t]
    \centering
    \captionof{table}{\textbf{3D pose estimation of articulated
    objects~\cite{Fan2023CVPR} -- KPE Design Choices.} 
    See \Secref{sec:arctic} for details.} 
    \tablelabel{tab:arctic_design}
    \setlength{\tabcolsep}{12pt}
    \resizebox{\linewidth}{!}{
    \begin{tabular}{l c c c c c c}
        \toprule
        \multirow{2}{*}{\bf Method} & \multicolumn{2}{c}{\bf Object} &
        \multicolumn{2}{c}{\bf Contact} & \multicolumn{2}{c}{\bf Motion} \\
        \cmidrule(lr){2-3} \cmidrule(lr){4-5} \cmidrule(lr){6-7}
        & AAE & Success & CDev$_{ho}$ & MRRPE$_{ro}$ & MDev$_{ho}$ & Acc \\
        & ($^\circ$) $\downarrow$  & (\%) $\uparrow$ & ($mm$) $\downarrow$ & ($mm$) $\downarrow$ & ($mm$) $\downarrow$ & ($m/s^{2}$) $\downarrow$ \\
        \midrule
        \multicolumn{7}{l}{\it a) Validation Split (Sparse \camenc design
        choices)} \\
        \quad \methodname (sparse)             & \bf 5.9 & \bf 71.5 & 39.4     & \bf 29.7 & \bf 9.3 & \bf 8.7 \\
        \quad \methodname (sparse center only) & 7.2     & 66.8     & 44.5     & 33.1     & 10.9    & 10.2 \\
        \quad \methodname (sparse w/ glb feat) & 6.2     & 69.8     & \bf 37.2 & 30.6     & 9.7     & 9.2 \\ \midrule
        \multicolumn{7}{l}{\it b) Validation Split (Dense \camenc design choices)} \\ 
        \quad \methodname (dense)          & \bf 6.2 & \bf 71.4 & \bf 37.7 & \bf 29.3 & \bf 9.9 & \bf 9.4 \\
        \quad \methodname (dense w/ input) & 7.3     & 68.1     & 48.7     & 38.3     & 13.6    & 10.2 \\
        \midrule
        \multicolumn{7}{l}{\it c) Validation Split (Sparse \vs Dense)} \\ 
        \quad \methodname (sparse)           & \bf 5.9 & \bf 71.5 & \bf 39.4 & \bf 29.7 & \bf 9.3  & \bf 8.7 \\
        \quad \methodname (dense)            & 6.2     & 71.4     & 37.7     & 29.3     & 9.9      & 9.4 \\
        \bottomrule
    \end{tabular}}
\end{table}

\boldparagraph{Metrics}
We follow the evaluation protocol of the ARCTIC benchmark~\cite{Fan2023CVPR}
and report the following metrics: a) \textbf{AAE}: Average Absolute Error
between the ground truth degree of articulation and the prediction; b)
\textbf{Success}: percentage of predicted object vertices with L2 error $< 5\%$
of the diameter of the object; c) \textbf{CDev}: Contact Deviation is average
distance between the predicted in-contact hand-object vertices and ground
truth; d) \textbf{MRRPE}: Mean Relative-Root Position Error between the
predicted root keypoint location and the ground truth; e) \textbf{MDev}:
Between consecutive frames, Motion Deviation measures the disagreement in the
direction of the movement of in-contact hand-object vertices; f) \textbf{ACC}:
Acceleration error is the difference in the predicted acceleration of the
vertices and the ground truth. 

\begin{figure*}[t]
    \centering
    \includegraphics[width=0.49\linewidth]{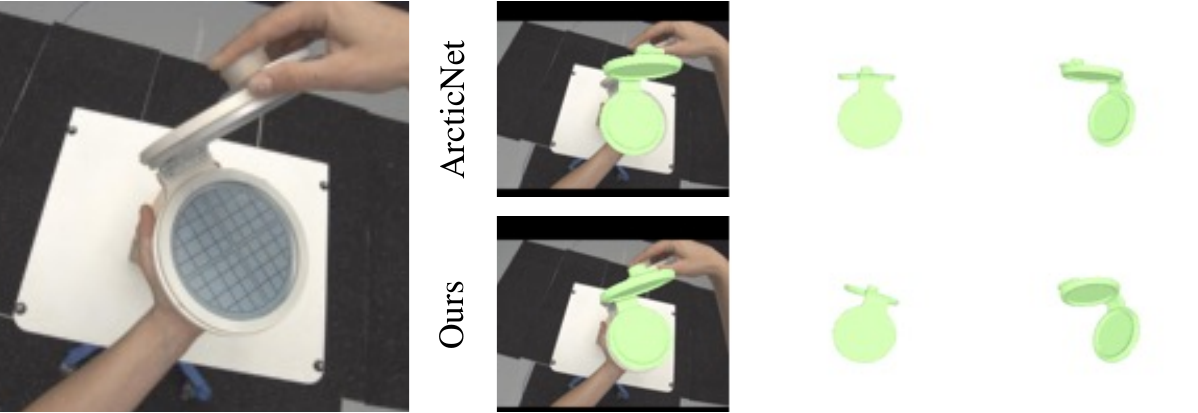}
    \hfill
    \includegraphics[width=0.49\linewidth]{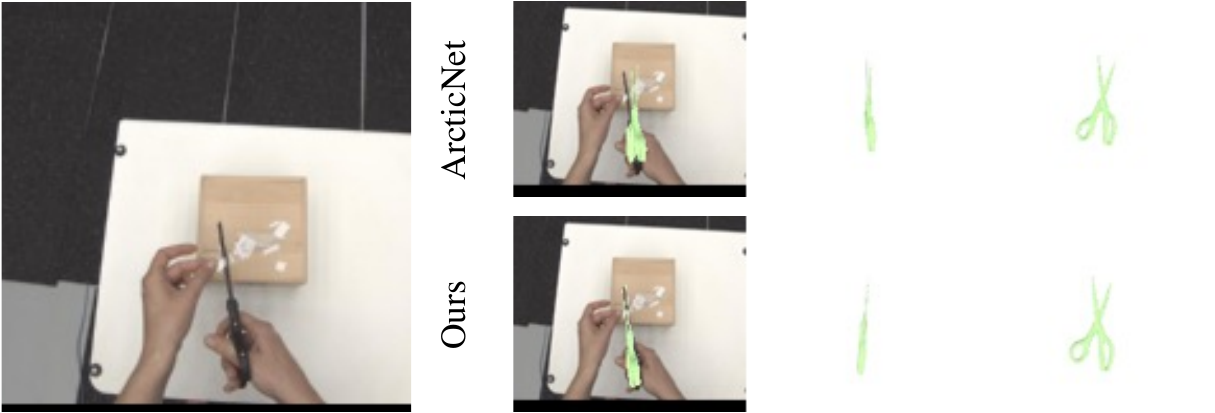}
    \caption{\textbf{3D pose visualizations on ARCTIC}. Our proposed modification of intrinsics-aware positional encoding (\methodname) improves over the ArcticNet-SF~\cite{Fan2023CVPR} model by predicting better 3D poses in interaction scenarios (note the difference in the articulation angle and global pose). For each image, we show the projection of the object mesh with the predicted pose on the image and from 2 different camera views.}
    \figlabel{fig:arctic_viz}
\end{figure*}

\boldparagraph{Results}
Adding \camenc to ArcticNet-SF leads to consistent improvements across the 3D
pose, contact and 4D motion estimation metrics across both validation
(\Tableref{tab:arctic_main}\,a) and test sets
(\Tableref{tab:arctic_main}\,b). Using full images as input, as done in
\cite{Fan2023CVPR}, obtains worse performance due to low resolution on the
object. Focusing on the object, either through masking or by inputting only a
crop, helps. KPE allows focusing on the object while also retaining the
location of the object in the visual field, leading to the best performance.
KPE made 2 design choices: a) use of sinusoids to encode, and b) use of crop's
location in the camera field of view rather than the 2D image. Both aspects are
important and KPE is better than Cam-Conv~\cite{facil2019cam} that represents
locations in camera's field of view but without sinusoids, and also just using
sinusoidal embeddings of the 2D location in image.

We also ablate design choices for where to inject sparse and dense KPE in
\tableref{tab:arctic_design}\,b and \tableref{tab:arctic_design}\,b. It is
beneficial to inject sparse KPE at earlier layers rather than at the very end
with global features. We also find that encoding both the scale and location of
the crop is better than just encoding the location. For dense KPE, inputting it
alongside the input is worse than using it alongside the $7\times 7$ spatial
feature map coming out from the ResNet50 backbne. 

While both sparse and dense versions of KPE improve performance, the sparse one
being slightly better (\tableref{tab:arctic_design}\,c).

For each image, we visualize (in \Figref{fig:arctic_viz}) the projection of the object mesh (assumed to be available) in the image using the predicted 7-dimensional pose (global rotation, camera translation \& articulation angle) along with renderings of the shape from two additional views. From the visualizations, we observe that \camenc leads to better angle of articulation and global rotation than ArcticNet-SF.

\subsection{Application 2: Dense Metric Depth Prediction~\cite{silberman2012indoor, bhat2023zoedepth}}
\seclabel{sec:depth}

\boldparagraph{Task} Given a single RGB image, the goal is to predict the
distance of each pixel to the camera center along the optical axis.

\boldparagraph{Method}
We adopt the ZoeDepth architecture~\cite{bhat2023zoedepth}. It consists of a
transformer-based MiDaS~\cite{Ranftl2022PAMI} backbone to estimate relative
depth, which is passed to a domain specific decoder along with multi-scale
image features from the backbone, to predict the metric depth for a specific
dataset. The model uses a discretized representation of 3D space (into a
volumetric grid, referred to as bins) to first output coarse depth (represented
as the center of the bin), which is then refined by an MLP in subsequent layers
(referred to as the MetricBins~\cite{bhat2023zoedepth} module). The model is
trained using the scale-invariant log loss (similar to~\cite{Bhat2022ECCV})
with $384 \times 512$ images without any cropping and scaling augmentations
during training. There are two variants of the model: a) ZoeD-X-N, trained from
scratch and b) ZoeD-M12-N, initialized from a MiDAS model pre-trained on $12$
datasets (also on $384 \times 512$ images).

\begin{table}[t]
  \centering
  \caption{
  \textbf{Dense metric depth prediction on NYU (Main Result)}. \camenc leads
  to improved depth estimation accuracy when trained and tested with cropping
  and scaling augmentations. \camenc outperforms Cam-Conv~\cite{facil2019cam},
  a past encoding specifically design for depth estimation. We see gains in
  both settings with and without pre-training of the MiDaS backbone.}
  \tablelabel{tab:depth_main}
  \setlength{\tabcolsep}{4.8pt}
  \resizebox{1.00\linewidth}{!}{
  \begin{tabular}{p{8cm} l l l}
      \toprule
      \bf Method & \bf REL $\downarrow$ & \bf RMSE $\downarrow$ & \bf $\log_{10}$ $\downarrow$ \\
      \cmidrule(lr){1-4}
      \multicolumn{4}{l}{\it a) No Pre-training for MiDaS backbone (KPE helps and is better than other choices)}\\
      \quad ZoeD-X-N & 0.093 & 0.360 & 0.042 \\
      \quad ZoeD-X-N + Cam-Conv~\cite{facil2019cam} & 0.098 & 0.382 & 0.042 \\
      \quad ZoeD-X-N + \camenc (dense) & \textbf{0.088} \gain{darkgreen}{-5.4} & \textbf{0.334} \gain{darkgreen}{-7.2} & \textbf{0.038} \gain{darkgreen}{-9.5} \\
      \midrule
      \multicolumn{4}{l}{\it b) MiDaS backbone pre-trained on 12 datasets (KPE helps and is better than other choices)} \\
      \quad ZoeD-M12-N & 0.088 & 0.337 & 0.040 \\
      \quad ZoeD-M12-N + Cam-Conv~\cite{facil2019cam} & 0.096 & 0.376 & 0.041 \\
      \quad ZoeD-M12-N + \camenc (dense) & \textbf{0.082} \gain{darkgreen}{-6.8} & \textbf{0.313} \gain{darkgreen}{-7.1} & \textbf{0.035} \gain{darkgreen}{-12.5} \\
      \bottomrule
  \end{tabular}}
\end{table}

\begin{table}[t]
  \centering
  \caption{\textbf{Dense metric depth prediction on NYU (KPE Design Choices)}.
  We conduct these comparisons in the setting where the MiDaS backbone is
  pre-trained on 12 datasets. See \secref{sec:depth} for more details.}
  \tablelabel{tab:depth_design}
  \setlength{\tabcolsep}{12.8pt}
  \resizebox{1.00\linewidth}{!}{
  \begin{tabular}{p{9cm} l l l}
      \toprule
      \bf Method & \bf REL $\downarrow$ & \bf RMSE $\downarrow$ & \bf $\log_{10}$ $\downarrow$ \\
      \cmidrule(lr){1-4}
      \multicolumn{4}{l}{\it a) Dense design choices} \\
      \quad ZoeD-M12-N + \camenc (dense, inside MiDaS backbone) & \textbf{0.082} & \textbf{0.313} & \textbf{0.035} \\
      \quad ZoeD-M12-N + \camenc (dense, after MiDaS backbone) & 0.085 & 0.323 & 0.038 \\
      \midrule
      \multicolumn{4}{l}{\it b) Sparse design choices} \\
      \quad ZoeD-M12-N + \camenc (sparse, inside MiDaS backbone) & \bf 0.083 & \bf 0.321 & \bf 0.037 \\
      \quad ZoeD-M12-N + \camenc (sparse, after MiDaS backbone) & 0.087 & 0.331 & 0.038 \\
      \midrule
      \multicolumn{4}{l}{\it c) Sparse \vs Dense} \\
      \quad ZoeD-M12-N + \camenc (dense) & \textbf{0.082} & \textbf{0.313} & \textbf{0.035}  \\
      \quad ZoeD-M12-N + \camenc (sparse) & 0.083 & 0.321 & 0.037 \\ 
      \bottomrule
  \end{tabular}}
\end{table}

\begin{figure*}[t]
    \centering
        \includegraphics[width=1.0\linewidth]{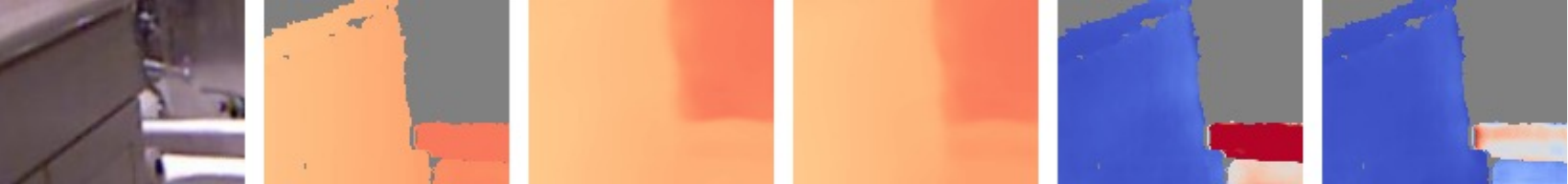} \\
        \includegraphics[width=1.0\linewidth]{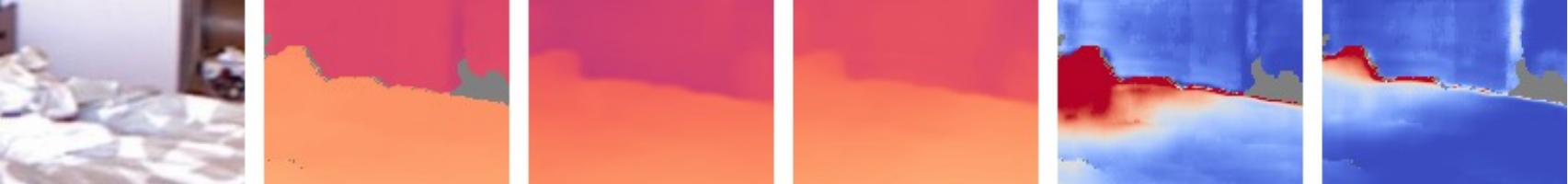} \\
        \scriptsize ~~~~~~Image \hfill GT Depth \hfill ZoeDepth \hfill Ours \hfill ZoeDepth $\Delta$ \hfill Ours $\Delta$~~~~~ \\
    \caption{{\bf Depth visualizations on NYU}. We compare the depth predicted by adding \methodname encoding to ZoeDepth~\cite{bhat2023zoedepth} with the base ZoeDepth model. We show the depth predictions along with the squared error $\Delta$ \wrt to ground truth depth (ranging from dark blue: low to dark red: high, invalid regions are indicated as grey). Our model predicts better depth as evident by lower $\Delta$ (lower intensity red areas).
    }
    \figlabel{depth_lowres_vis}              
\end{figure*}

\boldparagraph{Modifications}
We train and test on 3:4 aspect ratio image crops that are rescaled to $384
\times 512$ before being fed into the network. We do not change the
architecture or the loss function.  

\boldparagraph{Incorporating \camenc}
We experiment with both sparse and dense \camenc. We explore two design choices
for where to add \camenc: (1) with the relative depth \& $12\times16\times256$ image features predicted by the MiDaS model, before passing it to the MetricBins module, (2) with the downsampled $24\times32\times1024$ feature map before being processed by the BeiT~\cite{beit} module in MiDaS. We interpolate \camenc to a resolution of $12\times16$ for the former and $24 \times 32$ for the latter, in case of the dense variant. For the sparse variant, we broadcast it to the required resolution. %

\boldparagraph{Dataset} 
We use NYU Depth v2~\cite{silberman2012indoor}, a popular indoor scene dataset,
containing over 450+ unique scenes. We consider the same splits
as~\cite{bhat2023zoedepth}: 23K images for training, 654 images for testing and
compare against ZoeDepth~\cite{bhat2023zoedepth}. 

\boldparagraph{Metrics}
Following~\cite{bhat2023zoedepth}, we compute 3 metrics between the predicted metric depth ($\hat{d}_i$) and the ground truth ($d_i$): (1) \textbf{REL}: absolute relative error, $\frac{1}{M}\Sigma^{M}_{i=1}|d_i - \hat{d}_i|/d_i$, (2) \textbf{RMSE}: root mean squared error, $[\frac{1}{M}\Sigma^{M}_{i=1}|d_i - \hat{d}_i|^2]^{\frac{1}{2}}$, (3) \textbf{$\log_{10}$}: average $\log_{10}$ error, $\frac{1}{M}\Sigma^{M}_{i=1}|\log_{10}d_i - \log_{10}\hat{d}_i|$.

\boldparagraph{Results} 
Across both settings (pre-training and no pre-training), \camenc leads tom
improved performance over not using it, leading to relative improvements of
5.4\% -- 12.5\% across the different metrics and settings
(\tableref{tab:depth_main}\,a and b). We also compare to
Cam-Conv~\cite{facil2019cam}, a past encoding that has been used for single
image depth estimation. While Cam-Conv also encodes the location of pixels in
the camera's field of view, it doesn't encode them using sinusoidal embeddings
that are used in \camenc, and leads to worse performance.

In terms of design choices, we see more gains when adding \camenc to the downsampled
image features before it is passed to the BeiT~\cite{beit} module in the MiDaS
backbone than after the MiDaS backbone
(\tableref{tab:depth_design}\,a and b). Early infusion allows the transformer to
incorporate information about the crop location in the camera's field of view
in the multi-scale feature output. 
From~\Tableref{tab:depth_design}\,c, we find the dense variant of the encoding
to work better than the sparse variant. This is likely due to the dense
prediction nature of this task.

In~\Figref{depth_lowres_vis}, we visualize the predicted depth along with the
squared error $\Delta$ \wrt to ground truth (ranging from dark blue: low to
dark red: high). Our model predicts better depth as evident by lower $\Delta$
(lower intensity red areas). More visualizations are provided in supplementary.

\begin{table}[t]
    \centering
    \caption{\textbf{3D Object Detection (Main Results)}. We add \methodname to
    the Cube R-CNN~\cite{brazil2023omni3d} model and compare the AP scores at
    different IoUs. We see consistent improvements across all metrics.
    KPE helps and is better than other encodings. {\color{darkgreen} Green}
    numbers denote relative improvements over not using any encodings.}
    \tablelabel{tab:omni3d_main}
    \setlength{\tabcolsep}{4pt}
    \resizebox{\linewidth}{!}{
    \begin{tabular}{l l l l l}
        \toprule
        Method                                    & \apD{25}       & \apD{50}       & \apD{75}       & \apD{} \\
        \midrule
        Cube R-CNN                                & 70.17                                 & 43.83                                 & 14.23                                & 54.59 \\
        Cube R-CNN + Image location (dense)       & 71.97 & 45.03        & 15.16      & 56.86 \\
        Cube R-CNN + Image location (sparse)      & 72.08                                 & 45.93                                 & 15.32                                & 56.30\\
        Cube R-CNN + Cam-Conv~\cite{facil2019cam} & 72.62                                 & 46.51                                 & 14.59                                & 56.82 \\
        Cube R-CNN + \methodname (sparse)         & \textbf{73.10} \gain{darkgreen}{+4.2} & \textbf{47.51} \gain{darkgreen}{+8.4} & \textbf{16.58} \gain{darkgreen}{+16} & \textbf{57.54} \gain{darkgreen}{+5.4} \\
        \bottomrule
    \end{tabular}}
\end{table}

\begin{table}[t]
    \centering
    \caption{\textbf{3D Object Detection (KPE Design Choices)}. See
    \secref{sec:detection} for details.}
    \tablelabel{tab:omni3d_design}
    \setlength{\tabcolsep}{14pt}
    \resizebox{\linewidth}{!}{
    \begin{tabular}{l l l l l}
        \toprule
        Method                                    & \apD{25}       & \apD{50}       & \apD{75}       & \apD{} \\
        \midrule
        \multicolumn{5}{l}{\it Sparse \vs Dense KPE} \\
        $\quad$ Cube R-CNN + \methodname (dense) & 72.74 & 46.75 & 15.23 & 56.78 \\
        $\quad$ Cube R-CNN + \methodname (sparse) & \textbf{73.10} & \textbf{47.51} & \textbf{16.58} & \textbf{57.54} \\
        \bottomrule
    \end{tabular}}
\end{table}

\subsection{Application 3: 3D Object Detection on KITTI, nuScenes~\cite{brazil2023omni3d, geiger2012we, Caesar2020CVPR}}
\seclabel{sec:detection}

\boldparagraph{Task} The goal is to predict 3D
bounding boxes for objects, given a single image.

\begin{figure*}[t]
    \centering
    \begin{tabular}{c}
        \includegraphics[width=0.98\linewidth]{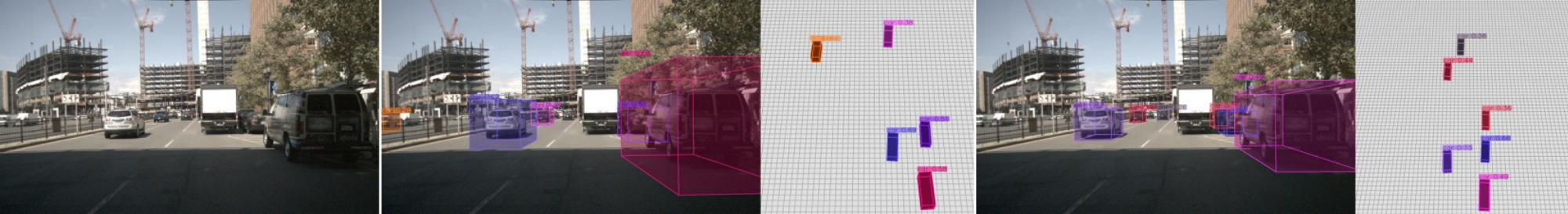} \\
        \includegraphics[width=0.98\linewidth]{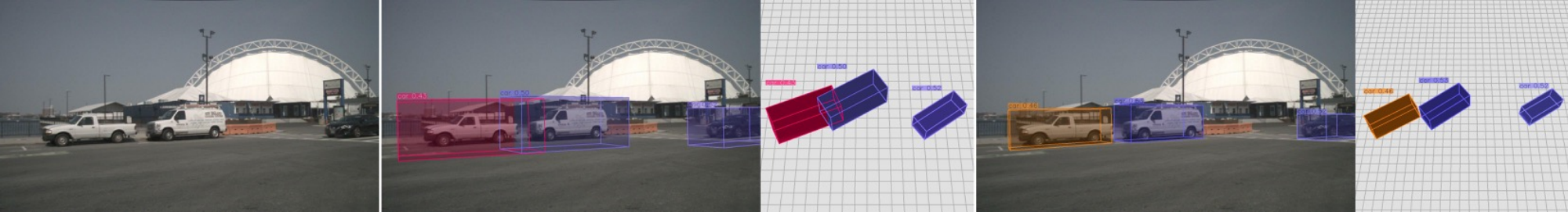} \\
        Image \hspace{0.22\linewidth} Cube R-CNN \hspace{0.25\linewidth} Ours \hspace{0.03\linewidth} \\
    \end{tabular}
    \caption{{\bf 3D Object Detection on nuScenes}. We show the 3D bounding box predictions on Cars category for Cube R-CNN~\cite{brazil2023omni3d} and our model (Cube R-CNN + \methodname), both in image space and in top-down view. Our model performs better, as evident by fewer collisions (\ie intersections between car bounding boxes) and missed detections. More visualizations on KITTI and nuScenes in supplementary.}
    \figlabel{nuscenes}              
\end{figure*}

\boldparagraph{Method}
We use the recent Cube R-CNN~\cite{brazil2023omni3d} model since it provides a generic framework for 3D detection, without any domain or object specific strategies. It consists of a convolutional backbone to extract features from an image, followed by a region proposal network~\cite{ren2015faster} to predict regions of interest (RoIs) (represented as 2D box proposals) used for detecting objects. The features from each detected object are passed to a decoder to predict a 3D bounding box.
The model is trained using 3D ground truth supervision for the 3D bounding box on multiple datasets jointly. Since each of these datasets are collected using different cameras, the authors propose to augment camera information with the ground truth depth used for training. This is referred to as virtual depth~\cite{brazil2023omni3d} and is orthogonal to our proposed \camenc encoding. Please refer to~\cite{brazil2023omni3d} for more details.

\boldparagraph{Modifications}
We use the same training protocol as~\cite{brazil2023omni3d}.

\boldparagraph{Incorporating \camenc}
The region proposal network in the backbone predicts 2D box proposals which are used for detecting objects. The positive box proposals with detected objects are then passed to the cube head to regress the 3D bounding box. We incorporate \camenc along with the features of the 2D box proposals before feeding to the cube head for 3D regression. Similar to \Secref{sec:arctic} and \Secref{sec:depth}, we explore both the sparse and the dense variant of our encoding. For the sparse variant, we broadcast the features to match the feature resolution whereas for the dense variant, we interpolate the features to the desired feature resolution.

\boldparagraph{Dataset}
We experiment on two commonly used datasets in 3D detection:
KITTI~\cite{geiger2012we} and nuScenes~\cite{Caesar2020CVPR}. These datasets capture urban outdoor scenes with a wide field of view, a setting in which perspective distortion is more prominent. We use the same splits as~\cite{brazil2023omni3d}: (1) KITTI: 2.8K training images \& 329 validation images, (2) nuScenes: 18K training \& 1.1K validation images, and focus on the Cars category. 
We consider the multi dataset setting where a single model is trained and tested on
KITTI and nuScenes. Since \camenc contains camera information, we hypothesize
that it should be effective in joint dataset training as well. In the joint
training on KITTI and nuScenes, we compare with the Cube-RCNN variant that does
not use virtual depth to disentangle the impact of virtual depth in
multi-camera setting. 

\boldparagraph{Metrics}
Following prior work~\cite{brazil2023omni3d}, we utilize 3D average-precision
(AP\textsubscript{3D}) as our metric to evaluate the 3D object detection
models. The predicted 3D cuboid is matched to the ground truth 3D cuboid by
computing a 3D intersection-over-union (IoU). We also report the
AP\textsubscript{3D} over a range of different IoU thresholds.

\boldparagraph{Results}
We observe consistent improvements in AP at different IoU
thresholds~\Tableref{tab:omni3d_main} with 4.2\% -- 16\% relative
improvement over not using any encoding. 

\tableref{tab:omni3d_design} compares \camenc's encoding of box location in the
camera field of view to just encoding the box location in the 2D image. We find
\camenc's encoding of the box's location in the camera's field of view to be
more effective. KPE is also more effective than Cam-Conv~\cite{facil2019cam}.
Sparse KPE works better than dense KPE.

We also visualize the 3D bounding box predictions on Cars category for Cube
R-CNN~\cite{brazil2023omni3d} \& our model (Cube R-CNN + \camenc)
in~\Figref{nuscenes}, both in image space \& in top-down view. Our model
performs better, as evident by fewer collisions (\ie intersections between car
bounding boxes) and missed detections. More visualizations on KITTI \& nuScenes
are in supplementary.

\section{Discussion}
\seclabel{discussion}

Our experiments provide several insights on three representative 3D from a single image tasks: 3D pose estimation of articulated objects in contact, dense metric depth prediction, 3D object detection, spanning dense \& sparse predictions and
convolutional \& transformer architectures. Specifically, across all settings, we found the use of \camenc, sparse or
dense, to improve performance. For applications involving sparse output (such
as 3D pose for objects), sparse \camenc worked better; while for applications
involving dense output (such as depth prediction), dense \camenc worked
better. \camenc is quite flexible and can be injected in the network at
different layers. The best location to inject \camenc is problem and method dependent. 
KPE has two important aspects: (a) use of sinusoids to encode, and (b) use of crop's
location in the camera field of view rather than the 2D image. Both aspects lead to KPE being better than Cam-Conv~\cite{facil2019cam} that represents locations in camera's field of view but without sinusoids, and uses sinusoidal encoding of the 2D location in image. At the same time, \camenc outperforms task specific prior encodings: Cam-Conv~\cite{facil2019cam} for depth prediction and 2D
image location for 3D object detection. Lastly
ablations on the task of 3D object pose estimation (\Secref{sec:arctic})
reveal that encoding both scale and location of the crop is better than just
encoding the location and \camenc is more effective than other ways of denoting
the location of the object, \eg via a masked image.

\boldparagraph{Limitations}
Our analysis only considers geometry of projection. It will be interesting to
assess if shading or other cues help alleviate the ambiguity. \camenc requires camera intrinsics which are typically present in image
metadata but may not always be available.
It would be useful to investigate if \camenc could be used with predicted intrinsics~\cite{jin2023perspective} in such cases. Jointly estimating instrinsics using appearance cues~\cite{jin2023perspective} in the image crop (\eg by adding a loss on the intrinsics) could also help with 3D predictions.
Other future directions include analyzing the impact of KPE on shallower networks, smaller receptive fields, predicting normals or full 3D shape from a single image.

\section{Conclusion}
We explore the ambiguity induced by perspective distortion in image crops. We provide an intuitive understanding of this ambiguity using parallelepipeds and propose an intrinsics-aware positional encoding (\camenc) that incorporates the location of the image crop in the field of the view of the camera. Experimental evaluation across three 3D from a single image tasks
(3D pose estimation of articulated objects in contact, dense metric depth
prediction and 3D object detection) reveals the effectiveness of \camenc.

\boldparagraph{Acknowledgements} We thank Matthew Chang, Shaowei Liu, Anand Bhattad and Kashyap Chitta for feedback on the draft, and David Forsyth for useful discussion. This material is based upon work supported by an NSF CAREER Award (IIS2143873), NASA (80NSSC21K1030), an NVIDIA Academic Hardware Grant, and the NCSA Delta System (supported by NSF OCI 2005572 and the State of Illinois).
\bibliographystyle{splncs04}
\bibliography{biblioShort, biblioLong, refs, refs_ambiguity}
\end{document}